\documentclass[runningheads]{llncs}
\usepackage[T1]{fontenc}
\usepackage{graphicx}
\usepackage{enumitem}
\usepackage{color}
\usepackage{tabularx}
\usepackage{makecell}
\usepackage{bbding}
\usepackage{makecell}
\usepackage{multirow}
\usepackage{multicol}
\usepackage{stfloats}
\usepackage{float}
\usepackage{afterpage}
\usepackage{placeins}
\usepackage{amsmath}
\usepackage{caption} 
\usepackage{subcaption}
\usepackage{pifont}
\usepackage{wasysym}

\usepackage{amssymb}

\begin{document}
\title{Mixed Text Recognition with Efficient Parameter Fine-Tuning and Transformer}
\titlerunning{PEFT-Transformer OCR for Mixed Text Recognition}
%
%
%

\author{Da Chang\inst{1}\orcidID{0009-0006-0216-4385} \and
Yu Li\inst{2}\orcidID{0009-0009-3734-9744} }


\institute{Shenzhen Institute of Advanced Technology, Chinese Academy of Sciences, China \and
The Institute of Technological Sciences, Wuhan University, China
\email{changda24@mails.ucas.ac.cn}\\
\email{yuuli2021@whu.edu.cn}}
\maketitle              
\begin{abstract}
With the rapid development of OCR technology, mixed-scene text recognition has become a key technical challenge. Although deep learning models have achieved significant results in specific scenarios, their generality and stability still need improvement, and the high demand for computing resources affects flexibility. To address these issues, this paper proposes DLoRA-TrOCR, a parameter-efficient hybrid text spotting method based on a pre-trained OCR Transformer. By embedding a weight-decomposed DoRA module in the image encoder and a LoRA module in the text decoder, this method can be efficiently fine-tuned on various downstream tasks. Our method requires no more than 0.7\% trainable parameters, not only accelerating the training efficiency but also significantly improving the recognition accuracy and cross-dataset generalization performance of the OCR system in mixed text scenes. Experiments show that our proposed DLoRA-TrOCR outperforms other parameter-efficient fine-tuning methods in recognizing complex scenes with mixed handwritten, printed, and street text, achieving a CER of 4.02 on the IAM dataset, a F1 score of 94.29 on the SROIE dataset, and a WAR of 86.70 on the STR Benchmark, reaching state-of-the-art performance.
\end{abstract}
\section{Introduction} 
\begin{figure}[h]
\centering
\includegraphics[width=0.5\textwidth]{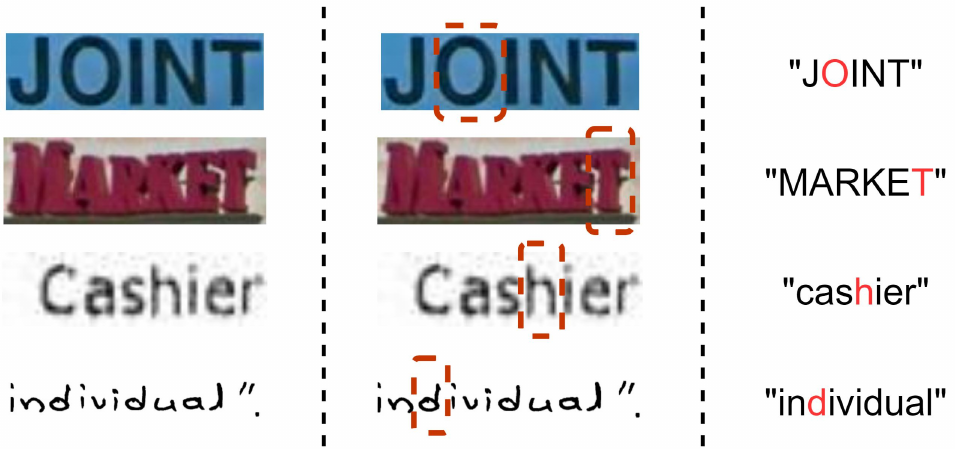}
\caption{Optical Character Recognition: Text Detection and Text Recognition.}
\label{fig1}
\end{figure}

OCR technology is used to classify optical patterns corresponding to letters, digits, or other characters in digital images \cite{Baviskar2021EfficientAP}. It comprises two main tasks: text detection, which locates text regions in an image, and text recognition, which extracts and interprets the text within those regions. As shown in Figure.\ref{fig1}, the red bounding boxes represent the text detection process, followed by the recognition of the content within these regions. Initially, OCR research focused on machine-printed text \cite{Memon2020HandwrittenOC}, primarily aiming to optimize text detection. Handwritten text recognition only gained significant attention in the late 20th century \cite{Raisi2020TextDA}, with English handwritten text posing notable challenges due to its diverse styles and character shapes. Additionally, the need for text recognition in complex multi-scene environments has grown. The ICDAR Robust Reading Competition introduced the first benchmark for scene text detection and recognition, \textit{ICDAR 2003} \cite{DBLP:conf/icdar/2003}. Over the next two decades, scene recognition benchmarks and methods, including deep neural networks, became more standardized and widely used \cite{Lin2019ReviewOS}, although the growing complexity of scene backgrounds continues to challenge text detection tasks in computer vision \cite{long2021scene}.

Recently, Transformer-based OCR methods for English text have greatly improved text recognition performance \cite{Fujitake_2024_WACV,li2023trocr}. These methods use the Transformer architecture to combine text detection and recognition into a single, end-to-end process. However, most models are designed for specific tasks, causing performance to drop significantly with mixed text types \cite{RABBY2021ADL}. Additionally, developing OCR technology for multi-scene and multi-type scenarios has been slowed by the high computational cost of training large models on extensive datasets \cite{Qin2024MultilingualLL}.

Our research focuses on broadening the application of the pre-training fine-tuning paradigm in OCR. To develop a versatile multi-scene text recognition model, we created a dataset that combines handwritten, printed, and scene text. We leveraged the pre-trained Transformer-based OCR model, TrOCR, using its weights trained on synthetic handwritten and printed text, and fine-tuned it on our mixed dataset. By incorporating parameter-efficient fine-tuning (PEFT) techniques, specifically applying DoRA to the encoder and LoRA to the decoder, we significantly reduced the number of trainable parameters while enhancing performance on the mixed dataset. Experimental results demonstrated exceptional performance in handling complex scenarios involving mixed handwritten, printed, and scene text, all with a minimal number of trainable parameters.

Our key contributions can be summarized as follows:
\begin{enumerate}
\item We constructed a comprehensive dataset combining English handwritten, printed, and complex scene text to encompass diverse image formats. 
\item We pioneered the development of a mixed-text OCR baseline model and introduced a novel application of the PEFT method, effectively addressing the challenges of fine-tuning Transformer-based Vision-Language models. 
\item We conducted extensive experiments on benchmark datasets for handwritten, printed, and scene text recognition tasks, validating the effectiveness and efficiency of our proposed approach. 
\end{enumerate}


\section{Related Work}
\subsection{OCR Framework.}

OCR models typically use a pipeline with modules like Convolutional Neural Networks (CNNs) for feature extraction \cite{gu2018recent} and Long Short Term Memory (LSTM) networks for sequence modeling \cite{yu2019review}. However, this modular design limits the ability to capture global information and long-range dependencies. The Transformer model, with its self-attention mechanism, has shown superior performance in sequence modeling tasks \cite{chen2021decision}. In OCR, it uses an encoder-decoder architecture to convert images to text. Sheng et al. \cite{Sheng2018NRTRAN} introduced NRTR, which uses a full Transformer architecture for encoding and decoding with minimal convolutional layers, proving the Transformer's effectiveness in text recognition. Yang et al. \cite{Yang2019AHR} presented SRACN, replacing LSTM with a Transformer decoder to enhance training efficiency and accuracy. Yu et al. \cite{Yu2020TowardsAS} proposed SRN, combining a Transformer encoder with a convolutional feature network. Li et al. \cite{li2023trocr} developed TrOCR, which utilizes a pre-trained Transformer for both image encoding and text decoding, bypassing the traditional CNN backbone.

\subsection{Parameter-Efficient Fine-Tuning Method.}
Parameter-Efficient Fine-Tuning (PEFT) \cite{pu2023empirical} aims to reduce the high cost of fine-tuning large models. This type of method achieves adaptation to downstream tasks by tuning a relatively small subset of parameters, rather than tuning all parameters of the model as in traditional methods. This strategy can significantly reduce computational cost and storage requirements compared to the total parameter amount. 

Existing PEFT methods can be mainly divided into two categories: addition-based methods and reparameterization-based methods.Our research focuses on the second category of reparameterization-based methods, which are implemented by converting the adaptive parameters in the optimization process into a parametrically efficient form. The representative of this type of method is Low-Rank Adaptation(LoRA) \cite{Hu2021LoRALA}, which assumes that changes in weights during model tuning have low intrinsic rank. Based on this assumption, LoRA proposes a low-order decomposition optimization approximation method for the variation of the original weight matrix in the self-attention module. Subsequently, Liu et al. introduced the idea of weight normalization based on LoRA, decomposed the pre-training weight into two components, amplitude and direction, and used LoRA to update the direction component. This method, called DoRA (Weight-Decomposed Low-Rank Adaptation) \cite{Liu2024DoRAWL}, enhances the learning capabilities and training stability of LoRA on some tasks without introducing any additional inference overhead.

\section{Approach}
Our proposed model is based on the TrOCR architecture and incorporates a pre-trained model with large-scale parameters to enhance generalization and robustness in multi-scene and multi-task OCR recognition. To address the challenge of efficiently fine-tuning large-scale Transformer parameters in OCR tasks, we employ the PEFT framework, which allows for effective fine-tuning of these parameters within the Transformer architecture. Specifically, we use the DoRA method to optimize the image encoder and the LoRA method to enhance the text decoder. This approach not only significantly reduces the number of trainable parameters but also improves the overall performance and inference efficiency of the encoder-decoder module.

\begin{figure}[!t]
    \centering
    \begin{subfigure}[b]{1.0\textwidth}
        \includegraphics[width=1.0\textwidth]{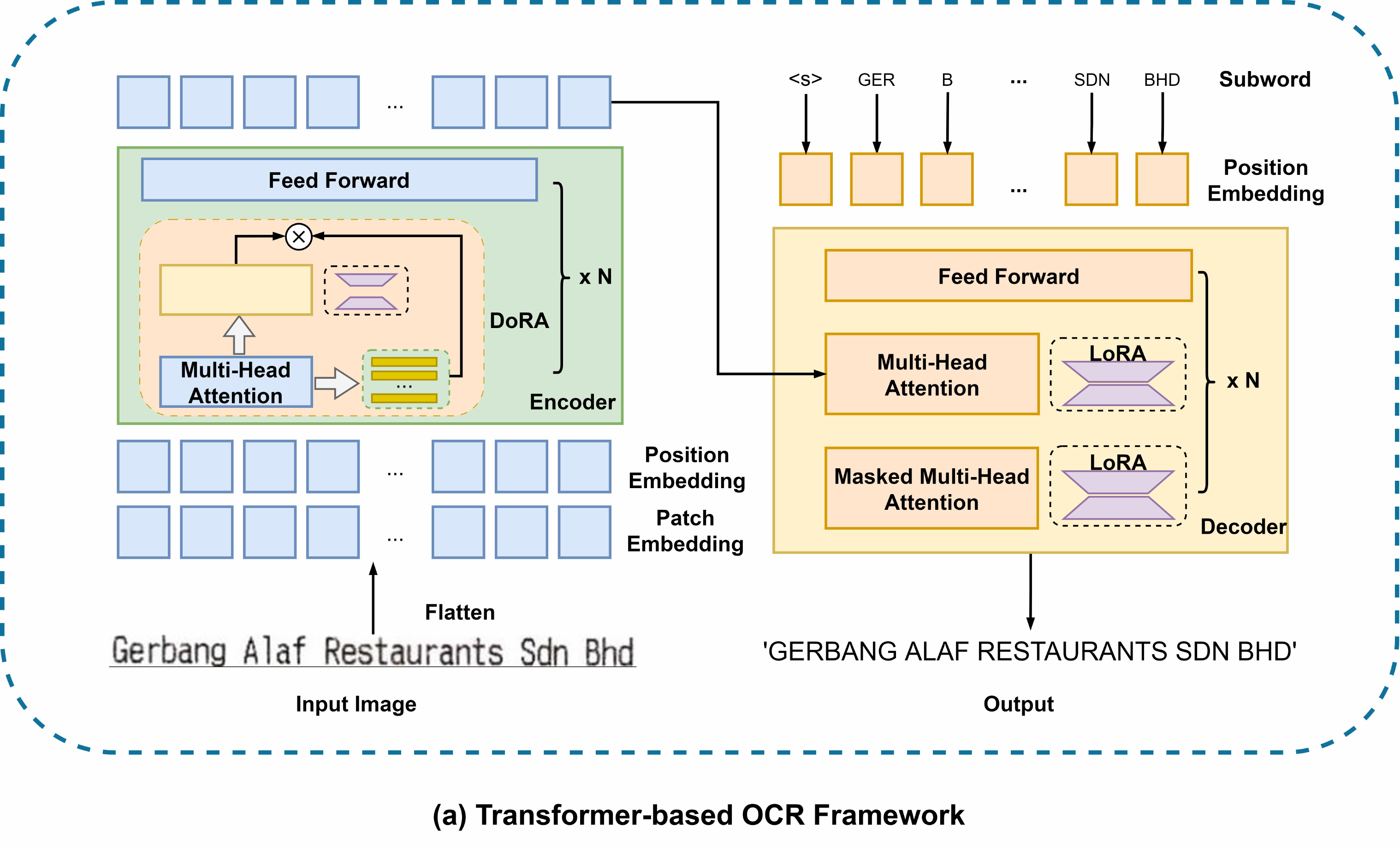}
        \label{fig:image1}
    \end{subfigure}
    \hfill
    \begin{subfigure}[b]{1.0\textwidth}
        \includegraphics[width=1.0\textwidth]{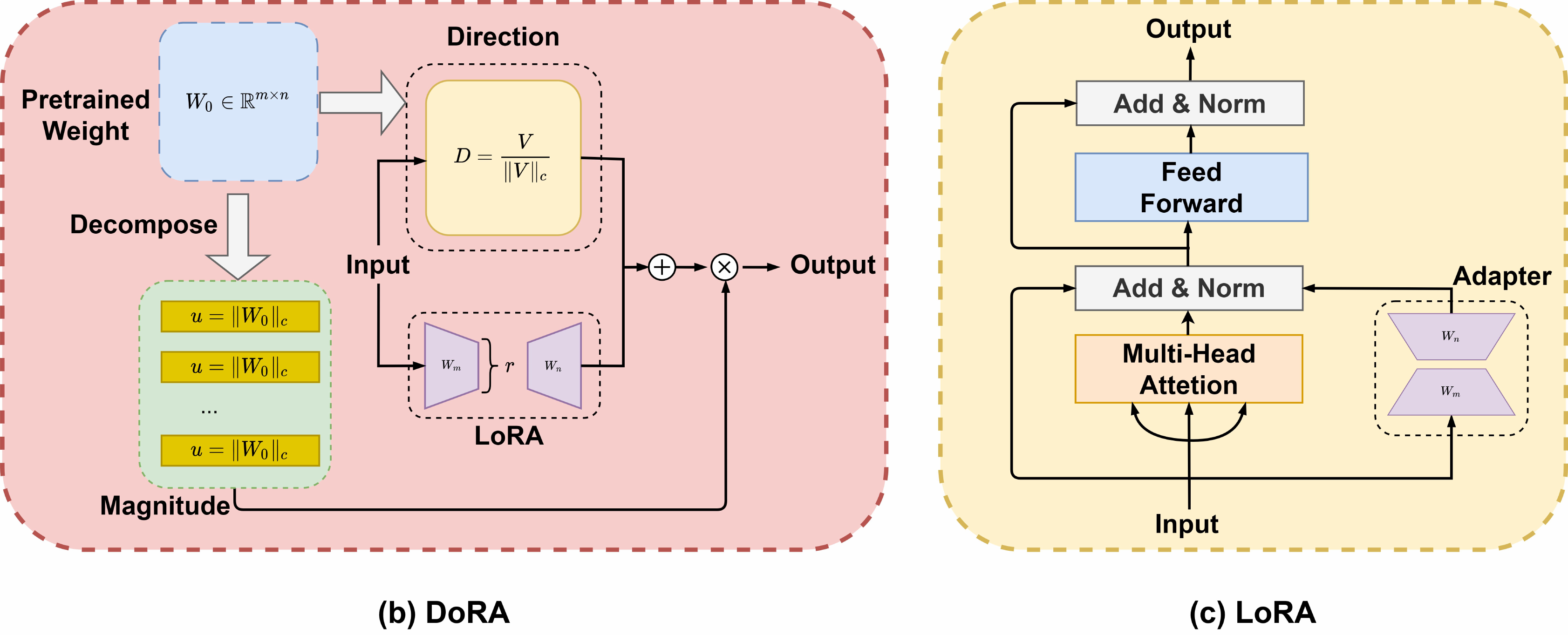}
        \label{fig:image2}
    \end{subfigure}
    \caption{Figure 2.(a) shows the transformer-based architecture, where the encoder-decoder model consists of a pre-trained image transformer as the encoder and a pre-trained text transformer as the decoder. The model is pre-trained in two stages on a synthetic dataset containing millions of handwritten and printed texts. The overall framework of our model is based on TrOCR \cite{li2023trocr}. Figures 2.(b) and 2.(c) illustrate the schematic diagrams of the DoRA and LoRA methods, respectively. As shown, DoRA decomposes the pre-trained weights into directions and magnitudes, updating the increments using LoRA in the direction before computing with magnitudes. The LoRA method approximates the weight update using a low-rank matrix.}
    \label{fig:images}
\end{figure}

\subsection{Transformer-based OCR}

\textbf{Architecture of OCR Transformer.}
The Transformer architecture we adopted incorporates an image Transformer for visual feature extraction and a text Transformer for language modeling. The overall model structure is depicted in Figure.\ref{fig:images}(a). Our model retains the original Transformer encoder-decoder structure, maintaining consistency in processing. The encoder focuses on obtaining representations of image patches, while the decoder produces word sequences guided by visual features and previous predictions.

\textbf{Image Encoder.}
The image encoder utilized is derived from the ViT model \cite{Dosovitskiy2020AnII}, designed to process an input image $x_{\mathrm{img}} \in \mathbb{R}^{3 \times H_0 \times W_0}$ by resizing it to a standard dimension $(H, W)$. The resized image is then segmented into $N$ patches of size $(P, P)$. Each patch is flattened into a vector and linearly projected into an embedding space of dimension $D$. To retain positional context, absolute positional embeddings are incorporated into each patch's embedding. These embeddings are subsequently processed by a sequence of Transformer encoder layers, where each layer includes a self-attention mechanism and a feed-forward network. To enhance training stability and maintain gradient flow, normalization and residual connections are applied. In the self-attention mechanism, similarities between queries and keys are computed to weight the aggregation of values, forming the output. The feed-forward network generally comprises two dense layers with a nonlinear activation function (e.g., ReLU or Swish \cite{10335476}) in between. Multiple attention heads, each with unique parameters, enable the model to learn from different representation subspaces effectively.

\textbf{Text Decoder.}
The text decoder is based on the RoBERTa model \cite{liu2019roberta}. Unlike the image encoder, it integrates an encoder-decoder attention layer that follows the multi-head self-attention layer, allowing it to focus on the encoder's outputs. In this attention layer, keys and values are derived from the encoder's outputs, while queries are generated from the decoder's inputs. The encoder-decoder attention mechanism calculates the similarity between queries and keys, using these to weight the values accordingly. Despite being initialized from an encoder-only architecture, the decoder employs autoregressive constraints during training. This approach involves predicting the next word using only the partially generated output and the prior input sequence, without referencing future target tokens.

\subsection{Parameter-efficient Fine-tuning}
While Transformer models excel in task-specific OCR, their massive scale—often comprising billions of parameters—renders fine-tuning and deployment in complex scenarios highly resource-intensive. To address this challenge, we employ Parameter-Efficient Fine-Tuning (PEFT) methods, which can drastically reduce the number of parameters needed for fine-tuning while preserving model performance. Specifically, we utilize two PEFT techniques: DoRA for fine-tuning the image encoder and LoRA for fine-tuning the text decoder.

\textbf{LoRA.} LoRA is a lightweight fine-tuning technology that achieves efficient fine-tuning by adding two small matrices to the key layers of the pre-trained model. Given the original weight matrix $W_0 \in \mathbb{R}^{m \times n}$ of a pre-trained model layer, LoRA reparameterizes it as: 

\begin{equation}
    W = W_0 + BA ,
\end{equation}

where $B \in \mathbb{R}^{m \times r}$ and $A \in \mathbb{R}^{r \times n}$ are two small matrices, known as LoRA weights, which are learned during fine-tuning. The hyperparameter $r$ determines the rank of these LoRA weights. When $r \ll \min(m, n)$, the number of parameters required for the LoRA weights is substantially smaller than that of the original weight matrix. This low-rank update facilitates efficient fine-tuning of pre-trained models. During inference, the fine-tuned LoRA weights are combined with the original model weights to produce the final inference weights. Since only two small matrices need to be stored, LoRA significantly reduces the additional parameter storage required for fine-tuning. In our text decoder, we apply LoRA to the query, key, and value projection matrices of the self-attention modules in both the decoder and the encoder-decoder, as well as to the output linear layer.

\textbf{DoRA.}  DoRA builds upon LoRA technology by decomposing the pre-trained weights into two components during fine-tuning: a magnitude vector $\mathbf{u} \in \mathbb{R}^{1 \times n}$ and a direction matrix $V \in \mathbb{R}^{m \times n}$. LoRA is then utilized for efficient directional updates, allowing the incremental update to be expressed as:

\begin{equation} \Delta W = \mathbf{u} \frac{V+\Delta V}{|V+\Delta V|_c} = |W_0|_c \frac{W_0+BA}{|W_0+BA|_c}. \end{equation}

Notably, the magnitude vector continues to use the original $|W_0|_c$ obtained from the pre-trained weight decomposition, where $|\cdot|_c$ denotes column-wise normalization. DoRA specifically targets the weight matrices in the multi-head attention mechanism of each Transformer encoder block. 

Specifically, LoRA and DoRA are employed to reconstruct the multi-head attention mechanism within the Transformer block. The incremental update for each attention head can be expressed as follows:

\begin{equation}
    \operatorname{head}_i = \operatorname{self-att}({Q}_i,{K}_i,{V}_i),
\end{equation}

where the query, key, and value matrices in the self-attention mechanism are updated using either the LoRA or DoRA approaches. The updates for each approach are summarized in the Table \ref{tab:lora_dora_updates}.

\begin{table}[h!]
\centering
\caption{Summary of Weight Matrix Update in Multi-Head Attention module on LoRA and DoRA.}
\begin{tabular}{|>{\centering\arraybackslash}p{2cm}|>{\centering\arraybackslash}p{3cm}|>{\centering\arraybackslash}p{6cm}|}
\hline
\makecell[c]{Weight \\Matrix} & \makecell[c]{LoRA \\Update} & \makecell[c]{DoRA \\Update} \\
\hline
\makecell[c]{$\mathbf{Q}_i$} & \makecell[c]{$(W_{q,i} + B_{q,i}A_{q,i})\mathbf{x}$} & \makecell[c]{$\left(W_{q,i} + |W_{q,i}|_c \frac{W_{q,i} + B_{q,i}A_{q,i}}{|W_{q,i} + B_{q,i}A_{q,i}|_c}\right)\mathbf{x}$} \\
\makecell[c]{$\mathbf{K}_i$} & \makecell[c]{$(W_{k,i} + B_{k,i}A_{k,i})\mathbf{x}$} & \makecell[c]{$\left(W_{k,i} + |W_{k,i}|_c \frac{W_{k,i} + B_{k,i}A_{k,i}}{|W_{k,i} + B_{k,i}A_{k,i}|_c}\right)\mathbf{x}$} \\
\makecell[c]{$\mathbf{V}_i$} & \makecell[c]{$(W_{v,i} + B_{v,i}A_{v,i})\mathbf{x}$} & \makecell[c]{$\left(W_{v,i} + |W_{v,i}|_c \frac{W_{v,i} + B_{v,i}A_{v,i}}{|W_{v,i} + B_{v,i}A_{v,i}|_c}\right)\mathbf{x}$} \\
\hline
\end{tabular}
\label{tab:lora_dora_updates}
\end{table}

Table \ref{label1} provides a detailed overview of the computing resource consumption and parameter scale of the TrOCR-base model across different PEFT strategies. The table specifically compares the following scenarios: (1)Full LoRA or DoRA: Both the encoder and decoder use the LoRA or DoRA method. (2)Hybrid configuration: The encoder is fine-tuned with the DoRA method, while the decoder is fine-tuned with the LoRA method. (3)Full parameter fine-tuning: All parameters are fine-tuned without applying any PEFT methods.

\begin{table}
\centering
\caption{Optimization of computational resources, model trainable parameters and inference speed using LoRA, DoRA, and DLoRA methods, which load the pre-trained weights of the model using half-precision (fp16).}
\label{label1}
\begin{tabular}{|>{\centering\arraybackslash}p{2cm}|>{\centering\arraybackslash}p{1.4cm}|>{\centering\arraybackslash}p{1.4cm}|>{\centering\arraybackslash}p{1.4cm}|>{\centering\arraybackslash}p{1.4cm}|>{\centering\arraybackslash}p{1.4cm}|>{\centering\arraybackslash}p{2.4cm}|}
\hline
Model &  \makecell{PEFT \\ Method} & \makecell{GPU \\ Memory} &  \makecell{Trainable \\ Params} &  \makecell{Total \\ Params} &  \makecell{Trainable \\ Ratio} & \makecell{Inference \\ Speed} \\
\hline
TrOCR-base & - & 23.28 & 333.9M & 333.9M & 100\%  & 691.79 token/s \\
TrOCR-base & LoRA & 15.01 & 2.0M & 335.9M & 0.585\%  &666.26 token/s \\
TrOCR-base & DoRA & 16.44 & 2.1M & 335.9M & 0.614\%  &640.14 token/s \\
TrOCR-base & DLoRA & 15.30 & 2.0M & 335.9M & 0.594\%  &656.03 token/s\\
\hline
\end{tabular}
\end{table}

For models utilizing PEFT methods, half-precision (fp16) is employed to load and freeze the pre-trained weights. This strategy accounts for only 0.6\% of the trainable parameters, reducing GPU memory usage by 30\% with the same batch size. In fact, the precision of the floating-point numbers can be reduced even further, allowing a significant increase in batch size and improving training efficiency while maintaining the same memory usage. Through the low-rank adaptation technique, the model trainable parameters can be significantly reduced. Since most of the weight parameters are still frozen, the model will not overfit to a specific task.

\section{Experiments}
\subsection{Experiment Setup}
\subsubsection{Dataset.}
\textbf{Handwritten dataset:} From the IAM dataset \cite{laia2016}, which contains handwritten English text and is one of the most commonly used datasets in the field of handwritten text recognition. We use the partitioning of the Aachen dataset: the training set contains 6,482 lines of text in 747 forms, the validation dataset contains 966 lines of text in 115 forms, and the test dataset contains 2,915 lines of text in 336 forms. \textbf{Printed dataset:} From the task2 of SROIE dataset  \cite{Huang2019ICDAR2019CO}, which focuses on text recognition in receipt images. The SROIE training dataset and test dataset contain 626 and 361 receipt images respectively. After cropping and removing duplicate images, the training dataset contains 10,682 lines of text, and the test dataset contains 6,897 lines of text. \textbf{Street View Text datasets:} From multiple public benchmarks, including IIIT5K-Words (IIIT5K) \cite{Mishra2009SceneTR}, Street View Text(SVT) \cite{Wang2011EndtoendST}, ICDAR 2013 (IC13) \cite{Karatzas2013ICDAR2R}, ICDAR 2015(IC15) \cite{Karatzas2015ICDAR2C}, Street View Text-Perspective (SVTP) \cite{Phan2013RecognizingTW}, and CUTE80 (CUTE) \cite{Risnumawan2014ARA}. These datasets contain various scene text images with noise such as blur, occlusion, or low resolution. The training set contains 7573 lines of text, and the test set contains 11435 lines of text. The specific dataset situation is shown in the Table \ref{tab:datasets}.

\begin{table}[h]
\centering
\caption{Handwritten, printed, and street view text datasets. Dataset Size represents the number of characters. The numbers in parentheses represent the number of lines of text that have been cropped. "\checked" indicates that the image has been cropped; "$\times$" indicates that the image does not need to be cropped.}
\label{tab:datasets}
\begin{tabular}{|>{\centering\arraybackslash}p{2.5cm}|>{\centering\arraybackslash}p{2cm}|>{\centering\arraybackslash}p{2cm}|>{\centering\arraybackslash}p{2cm}|>{\centering\arraybackslash}p{1cm}|}
\hline
\multirow{2}{*}{Dataset Type} & \multicolumn{4}{c|}{Datasets} \\ \cline{2-5}
 &  Dataset &  Train Set Size &  Test Set Size & Crop \\
\hline
\makecell[c]{Handwritten} & IAM & 747 (6842) & 336 (2915) & \checked \\
\cline{1-5}
\makecell[c]{Printed} & \makecell{SROIE\\task2} & 626 (10682) & 361 (6897) & \checked \\
\cline{1-5}
\multirow{6}{*}{\makecell{Scene Text \\ Recognition  \\
Benchmarks}} & IC15 & 4468 & 3888 & $\times$ \\
\cline{2-5}
& IC13 & 848 & 2967 & $\times$ \\
\cline{2-5}
& III5K & 2000 & 3000 & $\times$ \\
\cline{2-5}
& SVT & 87 (257) & 249 (647) & \checked \\
\cline{2-5}
& SVTP & - & 645 & $\times$ \\
\cline{2-5}
& CUTE & - & 288 & $\times$ \\
\hline
\end{tabular}
\end{table}

To achieve better generalization on multi-scene tasks, we adopt hybrid measures on the dataset. Compared with the circular sampling method \cite{pmlr-v97-stickland19a} that may lead to errors in uneven data distribution. We adopt a direct and balanced data integration strategy, directly mixing the datasets of the above handwritten , printed and scene text. Such a mixed dataset eliminates the possibility of a single task dominating the training process, and promotes the comprehensive learning of the model for a variety of image types and corresponding text descriptions.
All subsequent comparison, combination and ablation experiments are trained and tested based on this mixed dataset.

\subsubsection{Evaluation Metric.}

We considered three evaluation metrics that are widely used in OCR systems: Character Error Rate (CER), F1 score, and Word Accuracy Rate (WAR). CER is evaluated at the character level and is case-sensitive, while the F1 score and Word Accuracy are evaluated at the word level and are case-insensitive.

CER is defined as the edit distance between the predicted text and the ground truth. It can be calculated as:
\begin{equation} \mathrm{CER} = \frac{S+I+D}{N} \end{equation}

where $S$ is the number of substitutions, $I$ is the number of insertions, $D$ is the number of deletions, and $N$ is the total number of characters in the ground truth.

Word Accuracy Rate, derived as $1-\mathrm{WER}$ (Word Error Rate), assesses the proportion of correctly recognized words, making it particularly useful for evaluating STR (Scene Text Recognition) benchmarks. WER and CER are similar benchmark scores, with the main difference being that WER calculates errors at the word level, whereas CER does so at the character level. The F1 score, as the harmonic mean of Precision and Recall, provides a balanced measure of a model's accuracy, where Precision reflects the accuracy of detected words and Recall reflects the completeness of word detection.

\subsubsection{Implementation Details.}
In the model configuration, we utilize the official pre-trained weights of the TrOCR model and employ the AdamW optimizer for training over 20 epochs with a batch size of 16. The initial learning rate is set to 1e-5 for full-parameter fine-tuning and 5e-5 for the PEFT method. For the PEFT method our hyperparameter Settings for both lora and dora are low-rank of 8, alpha of 32, and dropout of 0.1.

All comparative experiments are conducted using PyTorch on an Nvidia RTX 4090 GPU with 24 GB of memory. To accurately evaluate model performance and ensure the integrity of the training, validation, and testing processes, we split the mixed dataset into a training set and a validation set with a classic 9:1 ratio. Additionally, we preserve a separate test dataset before mixing, ensuring that the final performance evaluation of the model remains entirely independent of the training process.

\subsection{Experiment Result}
\subsubsection{Performance.}
We conducted a detailed comparative analysis between the proposed method and a variety of current state-of-the-art OCR technologies. The evaluation covers seven benchmark datasets, and adopts F1, CER and WAR as core evaluation indicators. Detailed results are detailed in Table \ref{tab:compare}.
The results indicate that DLoRA-TrOCR large model outperforms the compared methods across most evaluation datasets and metrics. Notably, it achieves a word accuracy rate of 88.07\% on the mixed dataset we constructed. Furthermore, it attains the highest scores for CER, F1, and WAR on the IAM and SROIE Task2 datasets, clearly demonstrating the effectiveness and advantages of combining pre-trained models with PEFT in a multi-task OCR framework.

\begin{table}[h]
\centering
\caption{Simultaneous evaluation of different models across IAM, SROIE, STR benchmarks, and a Mixed Dataset with average F1 scores, CERs, and WARs. S, B, and L represent the parameter scale of the model, which stands for small, base, and large respectively.}
\label{tab:compare}
\begin{tabular}{|c|ccc|ccc|>{\centering\arraybackslash}p{1.75cm}|>{\centering\arraybackslash}p{1.3cm}|}
\hline
\multirow{2}{*}{Model} & 
\multicolumn{3}{c|}{{IAM}} & 
\multicolumn{3}{c|}{{SROIE}} & 
\multicolumn{1}{c|}{{\makecell{STR\\Benchmark}}} & 
\multicolumn{1}{c|}{{\makecell{Mixed\\Dataset}}} \\
\cline{2-9}
& {F1$\uparrow$} & {CER$\downarrow$} & {WAR$\uparrow$} & 
{F1$\uparrow$} & {CER$\downarrow$} & {WAR$\uparrow$} & 
{WAR$\uparrow$} & {WAR$\uparrow$} \\
\hline
CRNN \cite{shi2016end}  & 66.30 & 10.8 & 65.87 & 80.92 & 5.54 & 84.47 & 79.55 & 76.63 \\
CLOVA OCR \cite{Baek2019WhatIW}  & 67.71 & 9.24 & 70.77 & 84.41 & 4.88 & 86.23 & 79.55 & 76.63 \\
SRN \cite{9156632}   & 77.36 & 9.11 & 72.12 & 87.33 & 4.91 & 88.66 & 82.71 & 81.16 \\
VisionLAN \cite{Wang2021FromTT}    & 75.68 & 7.99 & 77.65 & 92.22 & 2.74 & 90.07 & 81.69 & 83.11 \\
ABINet \cite{Fang2021ReadLH}   & 78.66 & 8.87 & 79.88 & 90.75 & 3.12 & 89.22 & 83.47 & 84.19 \\
Donut \cite{kim2022donut}   & 82.64 & 8.02 & 82.36 & 92.44 & 2.02 & 91.31 & 74.37 & 85.97 \\

\makecell{DLoRA-TrOCR(S)}  & 76.20 & 10.16 & 75.38 & 91.85 & 2.38 & 90.33 & 71.31 & 79.01 \\
\makecell{DLoRA-TrOCR(B)}   & 80.73 & 7.56 & 80.54 & 92.93 & 1.42 & 91.05 & 82.30 & 84.63 \\
\textbf{DLoRA-TrOCR(L)}   & \textbf{84.98} & \textbf{4.02} & \textbf{84.79} & \textbf{94.29} & \textbf{1.05} & \textbf{92.73} & \textbf{86.70} & \textbf{88.07} \\
\hline
\end{tabular}
\end{table}

\subsubsection{Result analysis.}

To quantify the impact of parameter fine-tuning, we evaluated the performance of various efficient fine-tuning methods under different configurations using the TrOCR-Base model, as shown in Table \ref{tab:performance}. The results indicate that DLoRA achieved the best overall performance, with the highest F1 scores and lowest Character Error Rates (CER) across all datasets (IAM, SROIE, and STR Benchmark), demonstrating its potential to enhance multi-task text OCR accuracy. Compared to full parameter fine-tuning, all PEFT methods (LoRA, DoRA, LDoRA, and DLoRA) showed improved results, particularly in reducing CER. This confirms the effectiveness of selectively applying different PEFT methods to the encoder and decoder components of the Vision-Language encoder-decoder architecture to maximize performance while maintaining parameter efficiency. And due to the introduction of the LoRA-type low-rank adaptation module, it has better interpretability than full parameter fine-tuning. 

\begin{table}[h]
\centering
\caption{Comparative experimental results for full parameter fine-tuning, encoder-decoder using LoRA or DoRA, encoder using LoRA and decoder using DoRA, encoder using DoRA and decoder using LoRA.}
\label{tab:performance}
\begin{tabular}{|c|cc|cc|cc|}
\hline
\multirow{2}{*}{{TrOCR-Base}} & 
\multicolumn{2}{c|}{{IAM}} & 
\multicolumn{2}{c|}{{SROIE}} & 
\multicolumn{2}{c|}{{\makecell{STR \\Benchmark}}} \\
\cline{2-7}
& {F1$\uparrow$} & {CER$\downarrow$} & 
{F1$\uparrow$} & {CER$\downarrow$} & 
{F1$\uparrow$} & {CER$\downarrow$} \\
\hline
Fine-tuning & 76.42 & 10.47 & 91.48 & 1.84 & 82.57 & 7.29 \\
LoRA & 80.71 & \textbf{7.47} & 92.74 & 1.47 & 83.31 & 6.88 \\
DoRA & 80.60 & 7.57 & 92.60 & 1.48 & 83.26 & 7.04 \\
LDoRA & 80.64 & 7.50 & 92.49 & 1.53 & 83.27 & 6.94 \\
\textbf{DLoRA} & \textbf{80.73} & 7.56 & \textbf{92.92} & \textbf{1.42} & \textbf{83.45} & \textbf{6.70} \\
\hline
\end{tabular}
\end{table}

\begin{figure}[!h]
\centering
\includegraphics[width=1.0\textwidth]{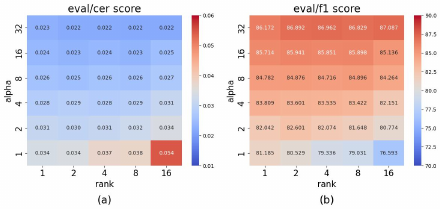}
\caption{The heatmap corresponds to low-rank values of 1, 2, 4, 8, and 16 alongside alpha values of 1, 2, 4, 8, 16, and 32. In Figure 3.(a), the color intensity represents the CER score on the validation set; a deeper blue indicates superior performance. Conversely, in Figure 3.(b), the colors reflect the F1 score on the validation set where more intense red hues signify better outcomes.}
\label{fig:sensitivity}
\end{figure}

We also performed a sensitivity analysis of the hyperparameters in DLoRA, specifically investigating the impact of low-rank size $r$ and alpha value on the attention mechanisms within both the decoder (query, key, value, and output) and encoder (query, key, and value). The critical parameters—alpha and rank—determine the scaling factor and dimensionality of the low-rank matrices in DLoRA, thereby directly influencing the magnitude of weight updates as well as the model's expressive capacity. The effectiveness of DLoRA is governed by the relationship $\mathrm{scale} = \frac{\alpha}{r}$.

Through a grid search illustrated in Figure \ref{fig:sensitivity}, we observed that smaller ranks combined with larger alpha values yield superior performance. Performance deteriorates when alpha is excessively low, irrespective of rank. As alpha increases while maintaining a fixed rank, performance improves up to a certain threshold before reaching saturation. Conversely, increasing rank at a constant alpha does not significantly enhance performance; this indicates that fine-tuning the alpha coefficient is more crucial for optimizing DLoRA's efficacy. It also indicates that DLoRA is more sensitive to alpha value than rank. Therefore, we strongly recommend paying special attention to the adjustment of the alpha value when using the peft method.

\subsection{Ablation Study.}

\begin{figure}[!t]
\centering
\includegraphics[width=1.0\textwidth]{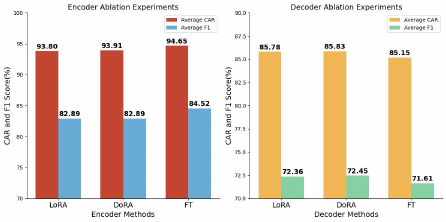}
\caption{The impact of DoRA, LoRA and Fine-Tune(FT) methods on the performance of our model's encoder and decoder modules, respectively. For comparison purposes, the metric used here is  Character Accuracy Rate, defined as ($1-\mathrm{CER})* 100\%$.}
\label{fig:endecoder}
\end{figure}

To demonstrate the optimal combination of PEFT functions, we conducted ablation experiments on the TrOCR-Base model. In these experiments, we fine-tuned either the encoder or the decoder while keeping the other frozen. Specifically, when the encoder was fine-tuned, the decoder's parameters were frozen, and vice versa. Figure \ref{fig:endecoder} presents the differences in average CER and F1 scores when using LoRA, DoRA, and full fine-tuning on both components.

The results of the ablation experiments reveal several insights: (1) In OCR tasks, applying the DoRA method to the image encoder alone shows a slight improvement over the LoRA method. This may be because DoRA effectively suppresses background noise during feature extraction through weight normalization, enhancing the model's robustness to image inputs. (2) Fine-tuning only the image encoder with all parameters performs better than fine-tuning both the image encoder and the text decoder together. This is likely because the pre-trained text decoder already has strong generalization capabilities, and fine-tuning it with all parameters on our relatively small mixed dataset could diminish this generalization. (3) Applying PEFT methods to the encoder proves to be far more effective than applying them to the decoder. 

We believe that the main reasons for the differences are as follows: 
(a) Diversity of Information: Image encoders need to process diverse information from various types of images, whereas text decoders handle relatively uniform text. This makes the fine-tuning of the image encoder more crucial for OCR tasks, while the impact on the text decoder is less significant. 
(b) Purpose of Feature Extraction: The primary role of the image encoder is to extract key features related to text content from images, while the text decoder predicts the next character based on these features and its previous predictions.

\section{Conclusion}

In this paper, we introduce the DoRA and LoRA methods from PEFT technology into the TrOCR model to enhance recognition accuracy and generalization across multiple datasets in complex mixed-text scenes. We apply DoRA to decompose the pre-training weights in the Encoder of the Transformer structure, allowing for stable processing of multi-text images by performing LoRA updates only on directional components. The Decoder uses LoRA to update the attention weights, improving the decoding capability for mixed text features. To test the model's robustness, we constructed a complex scene dataset that includes handwritten, printed, and scene text, and validated the superiority of our model in terms of accuracy and parameter efficiency through extensive comparative tests and ablation experiments. 
Our research presents new advancements in mixed-text OCR and provides valuable insights into optimizing parameters in existing models, laying a foundation for the development of more advanced, task-specific OCR techniques. By applying different PEFT methods to the encoder and decoder of Transformer models, we have improved performance across diverse OCR scenarios. This approach also offers guidance for other NLP tasks, such as information retrieval, highlighting its potential for broader applications in language processing.



\bibliographystyle{splncs04}
\bibliography{9753}
\end{document}